\documentclass[runningheads,orivec]{llncs}
\usepackage[width=122mm,left=12mm,paperwidth=146mm,height=193mm,top=12mm,paperheight=217mm]{geometry}

\usepackage{amsmath,amssymb} 
\usepackage{color}
\usepackage[breaklinks=true,colorlinks,citecolor=blue,urlcolor=blue,linkcolor=red,bookmarks=false]{hyperref}

\newcommand{\tabref}[1]{Tab. \ref{#1}}
\newcommand{\figref}[1]{Fig. \ref{#1}}

\def\ie{\emph{i.e.}}
\def\eg{\emph{e.g.}}
\def\etc{\emph{etc}}
\def\etal{{\em et al.~}}

\newcommand{\trb}[1]{\textbf{\textcolor{RedOrange}{#1}}}

\newcommand{\tbb}[1]{{\textbf{\textcolor{RoyalBlue}{#1}}}}

\usepackage{overpic}
\usepackage{booktabs}
\usepackage{multirow}
\usepackage{rotating}
\usepackage{wrapfig}
\usepackage{makecell}
\usepackage{cite}
\usepackage{authblk}

\usepackage[dvipsnames, table]{xcolor}

\RequirePackage{silence}
\graphicspath{{./Images/}{./Images_SM/}}

\begin{document}
\pagestyle{headings}
\mainmatter
\def\ECCVSubNumber{1615}  

\title{Gradient-Induced Co-Saliency Detection} 

\titlerunning{GICD}



 
\author{
 Zhao Zhang\inst{1} \orcidID{\href{https://orcid.org/0000-0002-1521-8163}{0000-0002-1521-8163}} 
 \and Wenda Jin\inst{2} \orcidID{\href{https://orcid.org/0000-0001-5964-4781}{0000-0001-5964-4781}} 
 \and \\ Jun Xu\inst{1} \orcidID{\href{https://orcid.org/0000-0002-1602-538X}{0000-0002-1602-538X}} 
 \and Ming-Ming Cheng\inst{1}  \orcidID{\href{https://orcid.org/0000-0001-5550-8758}{0000-0001-5550-8758}}
}
\authorrunning{Z. Zhang et al.}
\institute{TKLNDST, CS, Nankai University \and 
  College of Intelligence and Computing, Tianjin University \\
  \email{zzhang@mail.nankai.edu.cn; cmm@nankai.edu.cn}
}

\maketitle

\begin{abstract}
Co-saliency detection (Co-SOD) aims to segment the common salient foreground in a group of relevant images.
In this paper, inspired by human behavior, we propose a gradient-induced co-saliency detection (GICD) method.
We first abstract a consensus representation for a group of images in the embedding space; 
then,  
by comparing the single image with consensus representation,
we utilize the feedback gradient information to induce more attention to the discriminative co-salient features.
In addition, due to the lack of Co-SOD training data, 
we design a jigsaw training strategy, 
with which Co-SOD networks can be trained on general saliency datasets without extra pixel-level annotations.
To evaluate the performance of Co-SOD methods on discovering the co-salient object among multiple foregrounds,
we construct a challenging \textit{CoCA} dataset,
where each image contains at least one extraneous foreground along with the co-salient object.
Experiments demonstrate that our GICD achieves state-of-the-art performance.
Our codes and dataset are available at \url{https://mmcheng.net/gicd/}.
\end{abstract}

\section{Introduction}

Co-Saliency Detection (Co-SOD) aims to discover the common and salient objects 
by exploring the inherent connection of multiple relevant images.
It is a challenging computer vision task due to complex variations on the co-salient objects and backgrounds.
As a useful task for understanding correlations in multiple images, Co-SOD is widely employed as a pre-processing step for many vision tasks, such as weakly-supervised semantic segmentation~\cite{wei2016stc,zeng2019joint},
image surveillance~\cite{luo2015multi,gao2020trustful},
and video analysis~\cite{jerripothula2016cats,jerripothula2018efficient}, \etc.

Previous researches study the Co-SOD problem from different aspects~\cite{chen2010preattentive,li2019detecting,jiang2019unified}.
At the early stage, researchers explored the consistency among a group of relevant images using handcrafted features, \eg, SIFT~\cite{chang2011co,jerripothula2016cats}, color and texture~\cite{li2011co,fu2013cluster}, or multiple cues fusion~\cite{cao2014self}, \etc.
These shallow features are not discriminative enough to separate co-salient objects in real-world scenarios.
Recently, learning-based methods achieve encouraging Co-SOD performance by exploring the semantic connection within a group of images, 
via deep learning~\cite{wei2019deep,li2019detecting}, self-paced learning~\cite{zhang2016co,hsu2018unsupervised}, metric learning~\cite{han2017unified}, or graph learning~\cite{zheng2018feature,jiang2019unified}, \etc.
However, these methods suffer from the inherent discrepancy in features, due to varying viewpoints, appearance, and positions of the common objects.
How to better utilize the connections of relevant images is worth deeper investigation.
\begin{figure}[t]
	\centering
	\begin{overpic}[width=0.96\columnwidth]{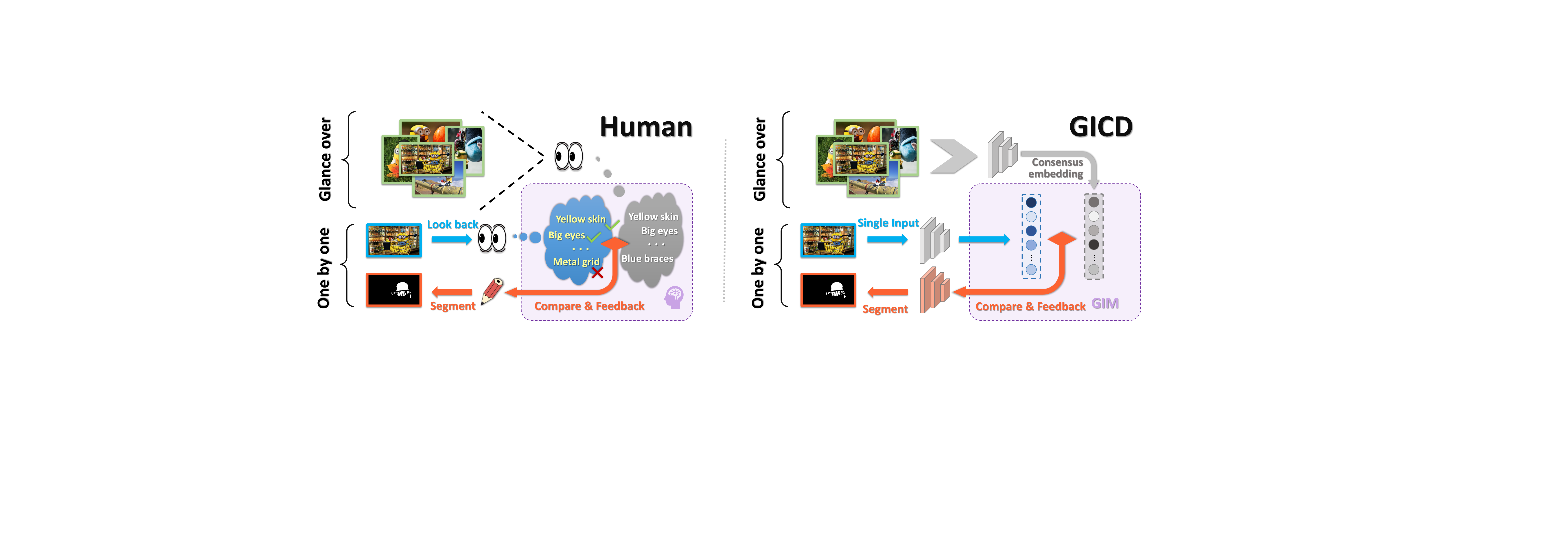}
	\end{overpic}
	\caption{
		\textbf{Human behavior inspired GICD}.
		GIM is the gradient inducing module.
	}
	\label{fig:human}
\end{figure}

How do humans segment co-salient objects from a group of images?
Generally, humans first browse the group of images, summarize the shared attributes of the co-salient objects with ``general
knowledge''~\cite{plaut2002graded}, and then segment the common objects in each image with these attributes.
This process is shown in \figref{fig:human}.
%
Inspired by human behavior, we design an end-to-end network with corresponding two stages.
%
To obtain the shared attributes of the common objects as humans do, we calculate the consensus representation of multiple relevant images in a high-dimensional space with a learned embedding network.
%
%
Once the consensus representation is obtained, 
for each image,
we propose a Gradient Inducing Module (GIM) to imitate the human behavior of comparing a specific scene with the consensus description to feedback matching information.

In GIM, the similarity between the single and consensus representations can be measured first.
As high-level convolutional kernels with different semantic awareness~\cite{zhou2016cam,selvaraju2017gradcam},
we can find out the kernels that are more 
related to the consensus representation
and enhance them to detect co-salient objects.
To this end,
by partially back-propagating,
we calculate the gradients of the similarity with respect to the top convolution layer as the feedback information.
High gradient values mean corresponding kernels have a positive impact on the similarity results;
thus, by assigning more weight to these kernels, 
the model will be induced to focus on the co-salient related features.
%
Moreover, to better discriminate the co-salient object in each level of the top-down decoder, 
we propose an Attention Retaining Module (ARM) to connect the corresponding encoder-decoder pairs of our model.
We call this two-stage framework with GIM and ARM as Gradient-Induced Co-saliency Detection (GICD) network.
Experiments on benchmark datasets demonstrate the advantages of our GICD over previous Co-SOD methods.

Without sufficient labels, existing Co-SOD networks~\cite{wei2019deep,li2019detecting,wang2020robust} are trained with semantic segmentation datasets, \eg, Microsoft COCO~\cite{lin2014microsoft}.
However, the annotated objects in segmentation datasets are not necessarily salient.
In this paper, we introduce a novel jigsaw strategy to extend existing salient object detection (SOD) datasets, without extra pixel-level annotating, for training Co-SOD networks.

In addition, to better evaluate the Co-SOD methods' ability of discovering co-salient object(s) among multiple foregrounds, most images in an evaluation dataset should contain at least one unrelated salient foreground except for the co-salient object(s). 
As can be seen in \figref{fig:dataset}, this is ignored by the current Co-SOD datasets~\cite{winn2005object,batra2010icoseg,zhang2016CoSal,fan2020taking}.
%
To alleviate the problem, we meticulously construct a more challenging dataset, named \textsl{Co}mmon \textsl{C}ategory \textsl{A}ggregation (\textit{CoCA}).
%



In summary, our major contributions are as follows:
\begin{itemize}
	\item \textbf{We propose a gradient-induced co-saliency detection (GICD) network for Co-SOD}.
	%
	Specifically, we propose a gradient inducing module (GIM) to pay more attention to the discriminative co-salient features, and an attention retaining module (ARM) to keep the attention during the top-down decoding.
	\item \textbf{We present a jigsaw strategy to train Co-SOD models on general SOD datasets without extra pixel-level annotations}, to alleviate the problem of lacking Co-SOD training data.
	\item \textbf{We construct a challenging \textit{CoCA} dataset with meticulous annotations}, providing practical scenarios to better evaluate Co-SOD methods.
	\item Experiments on the \textit{CoSal2015}~\cite{zhang2016CoSal} and our \textit{CoCA} datasets demonstrate that our GICD outperforms previous Co-SOD methods. 
	Extensive ablation studies validate the effectiveness of our contributions. 
\end{itemize}


\begin{figure}[t]
	\centering
	\begin{overpic}[width=.95\columnwidth]{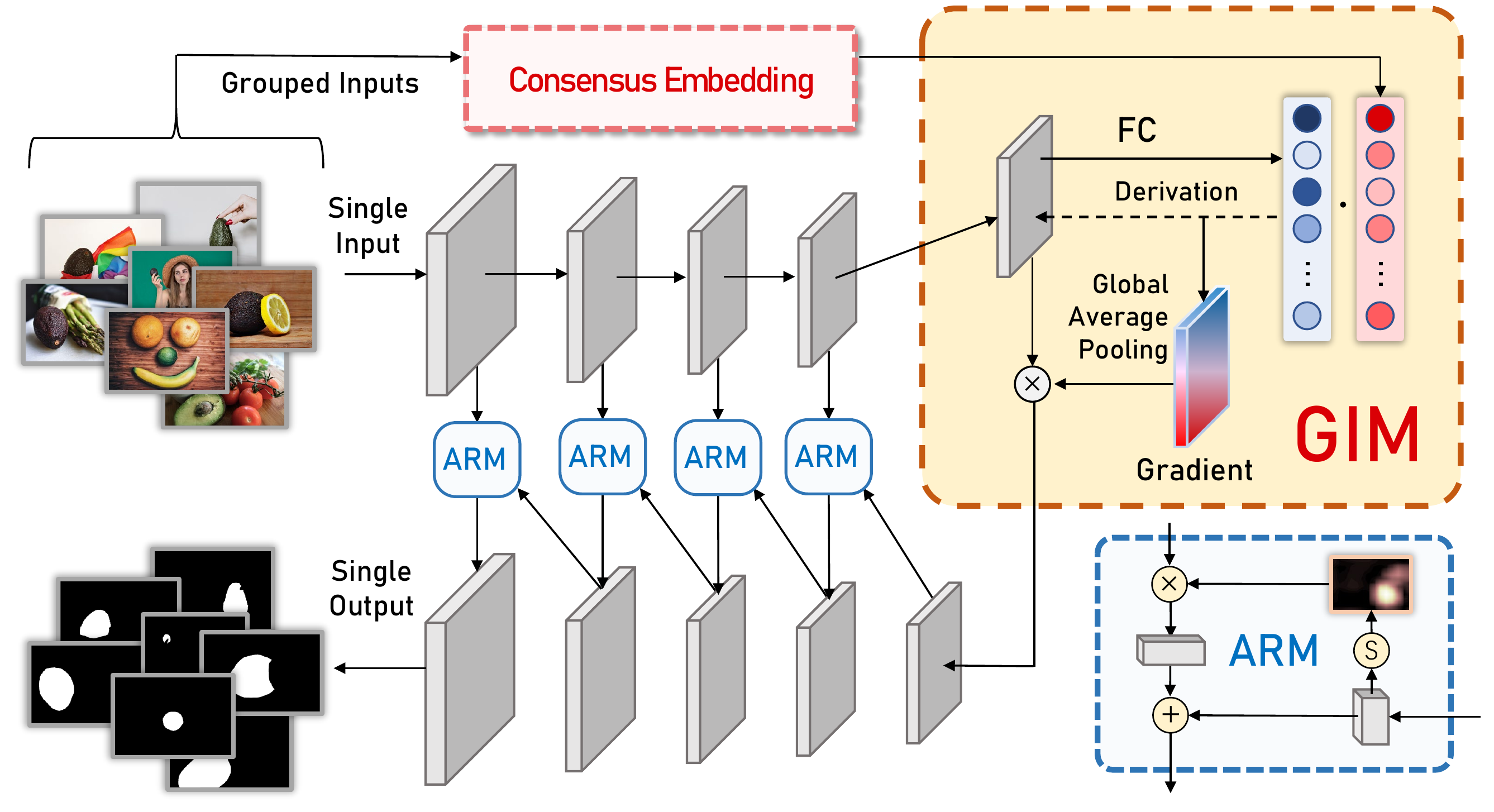}
	\end{overpic}
	\caption{\textbf{Pipeline of our Gradient-Induced Co-saliency Detection (GICD) method.}
		\textbf{GIM} denotes the Gradient Inducing Module, while \textbf{ARM} means the Attention Retaining Module.
		``$\bullet$'', ``\textcircled{$\times$}'', ``\textcircled{+}'', and ``\textcircled{s}'' represent the inner product, element-wise production, element-wise addition, and the sigmoid function, respectively.
	}
	\label{fig:pipeline}
\end{figure}

\section{Related Works}
\subsection{Co-Saliency Object Detection (Co-SOD)}
Different from traditional salient object detection (SOD) task~\cite{fan2018foreground,fan2020bbs,gao2020sod100k}, Co-SOD aims to automatically segment the common salient objects in a group of relevant images.
Early Co-SOD methods assume that the co-salient objects in multiple images share low-level consistency~\cite{zhang2016co}. 
For instance, Li~\etal\cite{li2011co} introduced a co-multi-layer graph by exploring color and texture properties.
Fu~\etal\cite{fu2013cluster} explored the contrast, spatial, and corresponding cues to enforce global association constraint by clustering.
 Cao~\etal\cite{cao2014self} integrated multiple saliency cues by a self-adaptive weighting manner. Tsai~\etal\cite{tsai2018image} extracted co-salient objects by solving an energy minimization problem over a graph.

Recently, many deep learning-based methods have been proposed to explore high-level features for the Co-SOD task~\cite{zhang2016CoSal,hsu2018unsupervised,zhang2019CSMG}.
These methods can be divided into two categories.
%
One is a natural deep extension from traditional low-level consistency.
It explores the high-level similarity to enhance the similar candidate regions among multiple images.
For example,
Zhang~\etal\cite{zhang2016CoSal} jointly investigated inter-group separability and intra-group consistency depending on high-level CNN features.
Hsu~\etal\cite{hsu2018unsupervised} proposed an unsupervised method by maximizing the similarity among multiple foregrounds and minimizing the similarity between foregrounds and backgrounds with graphical optimization.
Jiang~\etal\cite{jiang2019unified} explored the superpixel-level similarity by intra- and inter-graph learning using the graph convolution network.
Zhang~\etal\cite{zhang2019CSMG} proposed a mask-guided network to obtain coarse Co-SOD results and then refined the results by multi-label smoothing.
The second category of deep methods is based on joint feature extracting. 
They often extract the common feature for a group of images, and then fuse it with each single image feature.
For instance,
Wei~\etal~\cite{wei2019deep} learn a shared feature for every five images with a group learning branch, 
then concatenate the shared feature with every single feature to get the final prediction.
Li~\etal~\cite{li2019detecting} extend this idea with a sequence model to process variable length input.
Wang~\etal~\cite{wang2019robust} and Zha \etal\cite{wang2020robust} learn a category vector for an image group to concatenate with each spatial position of a single image feature on multiple levels.
%

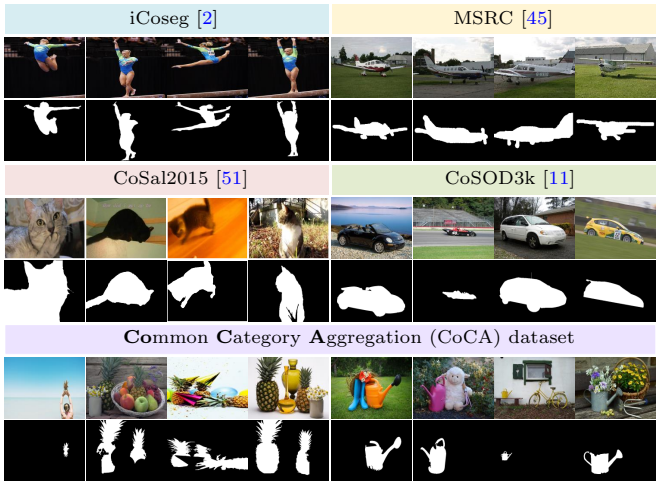
\begin{figure}[t]
	\centering
	\begin{overpic}[grid=false, width=0.72\columnwidth]{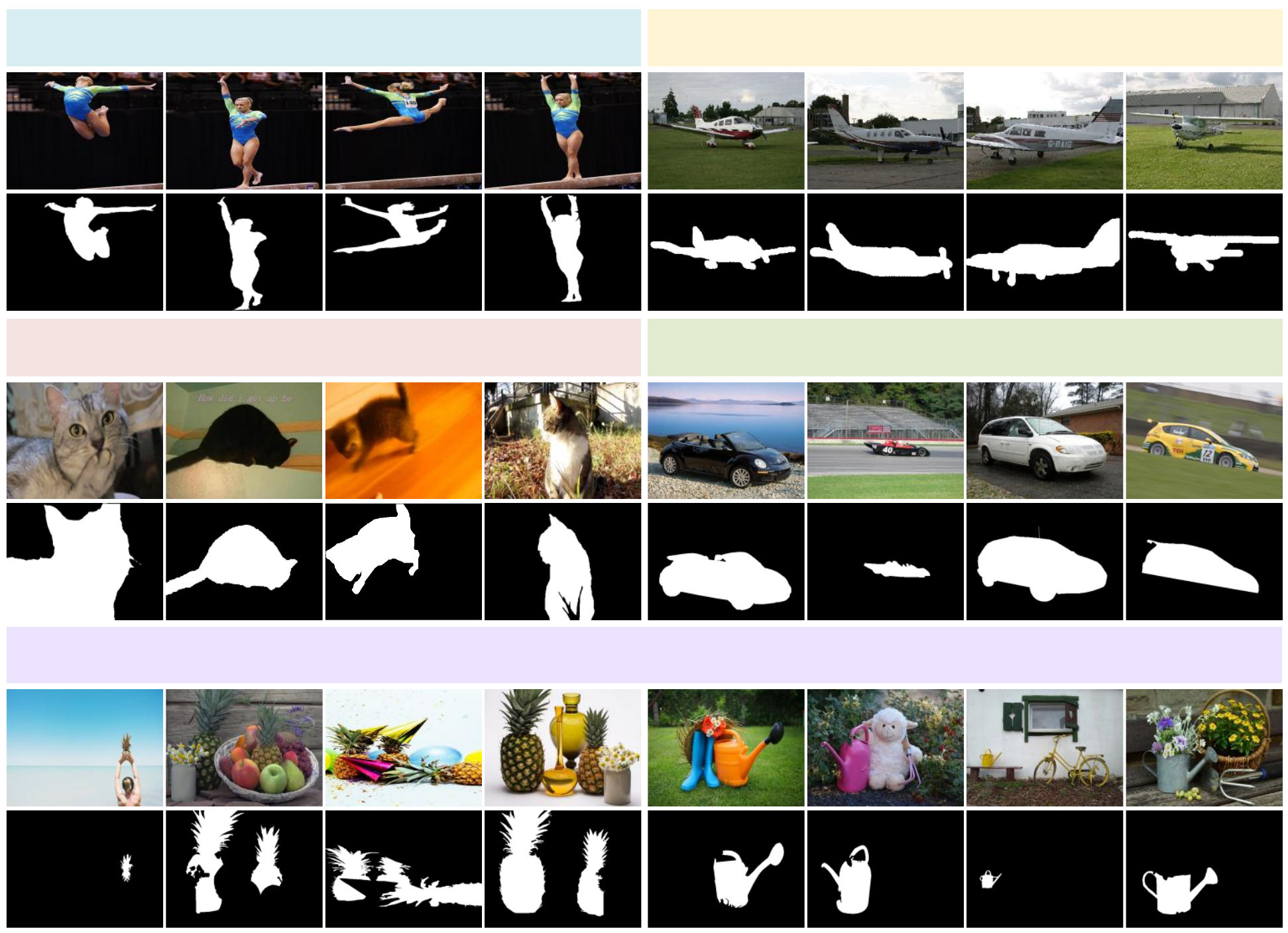}
	    \scriptsize
		\put(19.5,68.8){iCoseg~\cite{batra2010icoseg}}
		\put(68.8,68.8){MSRC~\cite{winn2005object}}
		\put(67.5,44.8){CoSOD3k~\cite{fan2020taking}}
		\put(17,44.8){CoSal2015~\cite{zhang2016CoSal}}
		\put(18.5,21){\textbf{Co}mmon \textbf{C}ategory \textbf{A}ggregation (CoCA) dataset}
	\end{overpic}
	\caption{
		\textbf{Examples of current popular datasets and our proposed \textit{CoCA} dataset}.
		In \textit{CoCA}, except for the co-salient object(s), each image contains at least one extraneous salient object, which enables the dataset to better evaluate the models' ability of discovering co-salient object(s) among multiple foregrounds.
	}
	\label{fig:dataset}
\end{figure}

\subsection{Co-SOD Datasets}
Current Co-SOD datasets include mainly \textit{MSRC}~\cite{winn2005object}, \textit{iCoseg}~\cite{batra2010icoseg}, 
\textit{CoSal2015}~\cite{zhang2016CoSal}, and \textit{CoSOD3k}~\cite{fan2020taking}, \etc.
In \figref{fig:dataset},
we show some examples of these datasets and our \textit{CoCA} dataset.
\textit{MSRC}~\cite{winn2005object} is mainly for recognizing objects from images.
In~\cite{fu2013cluster,zhang2016CoSal}, they select 233 images of seven groups from \textit{MSRC-v1} for evaluating detection accuracy.
%
%
\textit{iCoseg}~\cite{batra2010icoseg} contains 643 images of 38 groups in invariant scenes. 
In the above datasets, 
the co-salient objects are mostly the same in similar scenes, 
and consistent in appearance.
\textit{CoSal2015}~\cite{zhang2016CoSal} and \textit{CoSOD3k}~\cite{fan2020taking} are two large-scale datasets containing 2015 and 3316 images, respectively.
In the two datasets, some target objects belong to the same category differ greatly in appearance, which makes them more challenging datasets.
%
%
However, these above datasets are not well-designed for evaluating the Co-SOD algorithms
because they only have a single salient object in most images.
Taking the athlete of the \textit{iCoseg} in \figref{fig:dataset} as an example, 
although the athlete is co-salient in different images, 
these data can be easily processed by a SOD method
because there is no other extraneous salient foreground interference.
Although this awkwardness has been avoided in some groups in \textit{CoSal2015} and \textit{CoSOD3k}, it is not guaranteed in most cases.
As discovering the co-salient object(s) among multiple foregrounds is the primary pursuit of a Co-SOD method in real-world applications~\cite{zhang2018review},
to evaluate this ability better,
we construct a challenging \textsl{CoCA} dataset, 
where each image contains at least one extraneous salient object.

\section{Proposed Method}
\figref{fig:pipeline} shows the flowchart of our gradient-induced co-saliency detection (GICD) network.
Our backbone is the widely used Feature Pyramid Network (FPN)~\cite{lin2017feature}. 
For the Co-SOD task, we incorporate it with two proposed modules: the gradient inducing module (GIM), and the attention retaining module (ARM).
GICD detects co-salient objects in two stages.
It first receives a group of images as input for exploring a consensus representation in a high-dimensional space with a learned embedding network.
The representation describes the common patterns of the co-salient objects within the group.
Then, 
it turns back to segment the co-salient object(s) for each sample.
In this stage,
for inducing the attention of the model on co-salient regions,
we utilize GIM to enhance the features closely related to co-salient object by comparing single and consensus representation in the embedding space.
%
In order to retain the attention during the top-down decoding, we use ARM to connect each encoder-decoder pairs.
We train the GICD network with jigsaw training strategy, where the Co-SOD models can be trained on SOD dataset without extra pixel-level annotations.

\subsection{Learning Consensus Representation}
Given a group of images $\mathcal{I}=\{I_n\}_{n=1}^N$, to locate the co-salient object(s) in each image, we should first know what patterns the co-salient objects have based on prior knowledge.
To this end, we propose to learn a consensus representation with a pre-trained embedding network, for the co-salient objects of the image group $\mathcal{I}$.
%
Deep classifiers can be naturally utilized for representation learning~\cite{qi2018low}, where the prior knowledge of semantic attribute can be transformed from the parameters pre-trained on ImageNet~\cite{deng2009imagenet}.
In this case,
we employ a pre-trained classified network $\mathcal{F(\cdot)}$, such as VGG-16, as our embedding network by removing the softmax layer.
It first extracts the representation $e_n=\mathcal{F}(I_n)\in\mathbb{R}^d$ of each image $I_n$, 
where $d$ is the dimension of the last full connection layer.
The consensus representation $e^{\dag}$ can be calculated by $e^{\dag} = \mathtt{Softmax}\left(\sum_{n=1}^{N}e_n\right)$, to describe the common attributes of this image group.

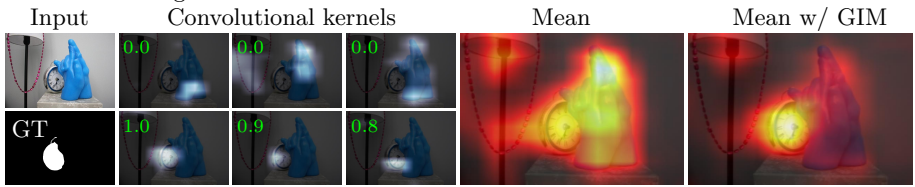
\begin{figure}[t]
	\centering
	\begin{overpic}[grid=false,width=1\columnwidth]{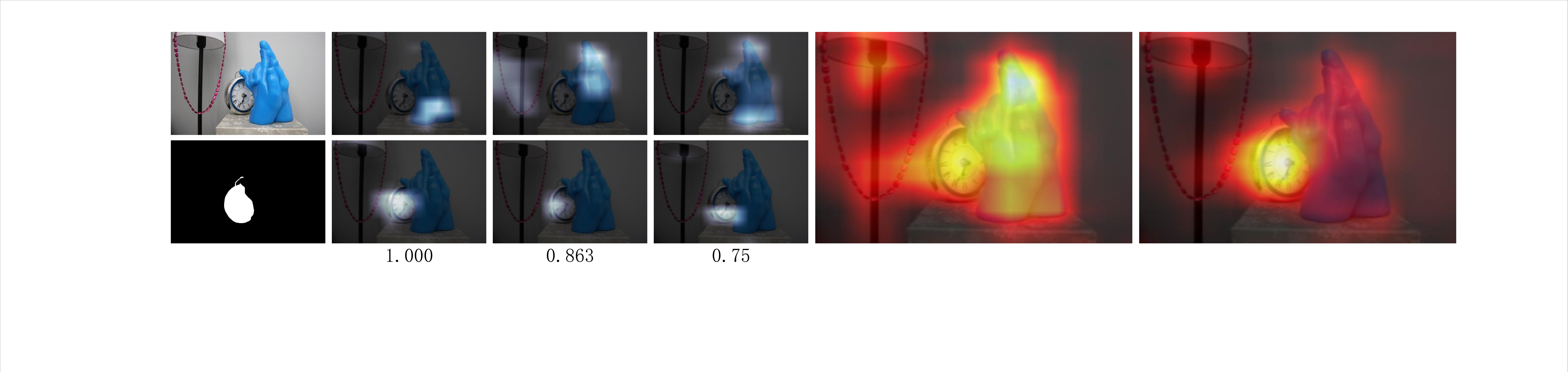}
		\put(18,17.5){Convolutional kernels}
		\put(58,17.5){Mean}
		\put(80,17.5){Mean w/ GIM}
		\put(3,17.5){Input}
		\put(1,5.5){\color{white}{GT}}
		\put(13.2,14.5){\scriptsize{\color{green}{0.0}}}
		\put(25.7,14.5){\scriptsize{\color{green}{0.0}}}
		\put(38.2,14.5){\scriptsize{\color{green}{0.0}}}
		\put(13.2,6){\scriptsize{\color{green}{1.0}}}
		\put(25.7,6){\scriptsize{\color{green}{0.9}}}
		\put(38.2,6){\scriptsize{\color{green}{0.8}}}
	\end{overpic}
	\caption{
		\textbf{Visualization of high-level features induced by GIM}.
		In the six small images on the left, 
		the kernels above are not sensitive to the target object, 
		while the kernels below are related to the target object.
		Their corresponding important values based on gradients are marked with \textbf{green} numbers on the top-left corners.
		The mean values of $F^5_n$ and induced $\tilde{F}^5_n$ are shown in the form of orange heat maps.
	}
	\label{fig:GIM}
\end{figure}

\subsection{Gradient Inducing Module}
After obtaining the consensus representation $e^{\dag}$ of the group $\mathcal{I}$,
for each image,
we focus on how to find the discriminative features that match the consensus description.
%
As demonstrated in \cite{zhou2016cam,selvaraju2017gradcam}, high-level convolutional layers naturally possess semantic-specific spatial information.
%
We denote the five convolutional feature blocks of  $\mathcal{F(\cdot)}$ as $\{F^1,F^2,\dots,F^5\}$.
In \figref{fig:GIM}, we show the feature maps of the last convolutional layer $F^5$.
The input image (1st column) contains a pocket watch and blue gloves, and the convolutional kernels focus on different regions (2nd to 4th columns).
If assigning more importance to these kernels which closely concern about the co-salient objects,
the model will tend to segment the co-salient objects (pocket watch) by decoding the induced features.
%
As indicated in~\cite{selvaraju2017gradcam},
the discriminability of features in neural networks can be measured by the gradient obtained by optimizing objectives.
Therefore, we propose a gradient inducing module (GIM) for enhancing the discriminative feature by exploring the feedback gradient information.
As the encoder of our FPN backbone shares the fixed parameters with the consensus embedding network, it can also embed each image into the same space as consensus representation $e^{\dag}$. 
%
For the extracted representation $e_n$ of the $n$-th image,
the similarity $c_n$ between $e_n$ and its consensus representation $e^{\dag}$ can be defined by inner product, \ie, $c_n = e_n^{\top} e^{\dag}$.
Then we compute the positive gradient $G_n$ flowing back into the last convolutional layer $F^5\in\mathbb{R}^{w\times h\times c}$ 
to select discriminative features in $F^5_n$, specifically,
\begin{equation}
	G_n = \text{ReLU}\left(\frac{\partial c_n}{\partial F^5_n}\right) \in\mathbb{R}^{w\times h\times c}.
\end{equation}
In this partial backpropagation, the positive gradient $G_n$ reflects the sensitivity of the corresponding position to the final similarity score; 
that is, increasing activation value with a larger gradient will make the specific representation $e_n$ more consistent with the consensus one $e^{\dag}$.
Therefore, the importance of a convolution kernel for a particular object can be measured by the mean of its feature gradients.
Specifically, the channel-wise importance values can be calculated by global average pooling (GAP), namely $w_n=\mathtt{GAP}(G_n)=\frac{1}{wh}\sum_i\sum_j G_n$, where $i=1,...,w$ and $j=1,...,h$.
Once obtaining the weight, we can induce the high-level feature $F^5_n$ by assigning the importance value to each kernel $\tilde{F}^5_n=F^5_n\otimes w_n$, where $\otimes$ denotes the element-wise production.
%
%
As shown in \figref{fig:GIM}, we visualize the mean heat-maps of $F^5_n$ and $\tilde{F}^5_n$.
without our GIM module, the kernels will averagely focus on both objects.
One can see that the kernels more relevant to the co-salient category have higher gradient weights (marked with green numbers), and the attention of the network has shifted to the co-salient object(s) after gradient inducing.

\subsection{Attention Retaining Module}
In GIM, the high-level features have been induced by the gradient.
However, top-down decoder is built upon the bottom-up backbone, and the induced high-level features will be gradually diluted when transmitted to lower layers.
To this end, we propose an attention retaining module (ARM) to connect the corresponding encoder-decoder pairs of our GICD network.
As shown in \figref{fig:pipeline}, for each ARM, the feature of encoder used for skip-connection is guided by the higher-level prediction.
Through top-down iterative reminding,
the network will focus the detail recovery of the co-salient regions without being interfered by other irrelevant objects.
We take the channel-wise mean of $\tilde{F}^5_n$ as the first low-resolution guide map $S_n^5$,
and reduce $\tilde{F}^5_n$ to feature $P_n^{5}$ containing 64 channels.
The decoding process with ARM is as follows:
\begin{equation}
\left\{
\begin{array}{l}
\tilde{F}^i_n = \left( S_n^{i+1} \right)\uparrow  \odot F_n^i \\
P_n^i = \mathcal{E}^i\left( \left(P_n^{i+1}\right)\uparrow + \mathcal{R}^i\left( \tilde{F}^i_n \right)  \right), \\
S_n^i = \mathcal{D}^i\left( P_n^i \right), \\
\end{array}
\right.
i\in\{4,3,2,1\},
\end{equation}
where $(\cdot)\uparrow$ is the up-sampling operation.
$\mathcal{R}^i(\cdot)$ consists of two convolutional layers, and reduces the enhanced features $\tilde{F}^i_n$ to 64 channels.
$\mathcal{E}^i(\cdot)$ is the corresponding two convolutional layers, with 64 kernels, in decoder.
$\mathcal{D}^i(\cdot)$ is applied for deep supervision, and outputs a prediction by two convolutional layers followed by a sigmoid layer.
The last $S^1_n$ is the final output.

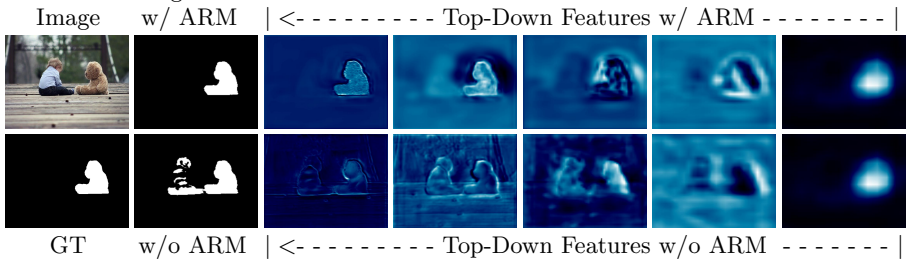
\begin{figure}[t]
	\centering
	\begin{overpic}[grid=false, width=1\columnwidth]{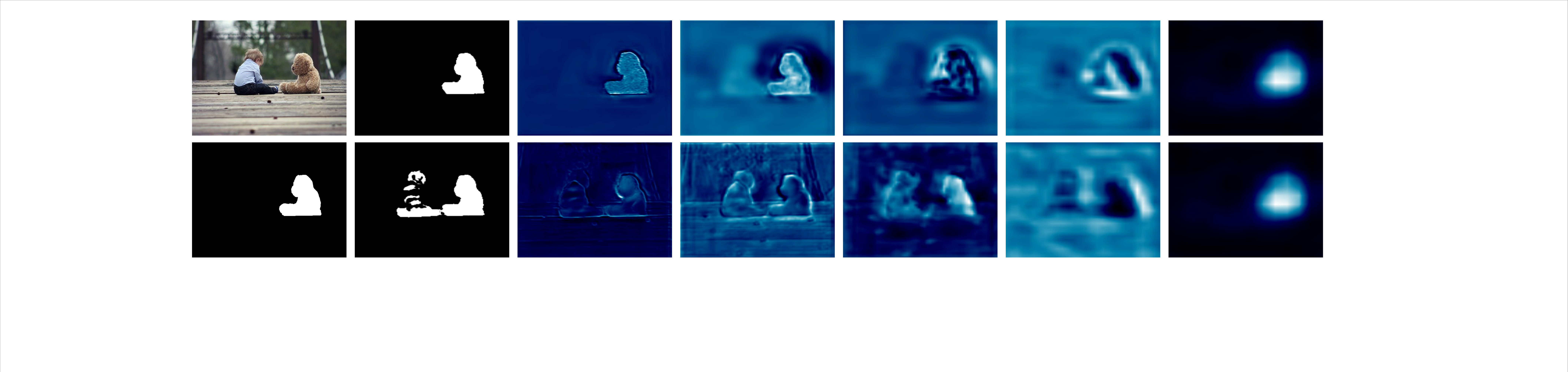}
		\put(4,22.5){Image}
		\put(15.5,22.5){w/ ARM}
		\put(29,22.5){$|<$-\ -\ -\ -\ -\ -\ -\ -\ -\ Top-Down Features w/\ ARM -\ -\ -\ -\ -\ -\ -\ -\ $|$}
		\put(5.5,-2){GT}
		\put(15.2,-2){w/o ARM}
		\put(29,-2){$|<$-\ -\ -\ -\ -\ -\ -\ -\ -\ Top-Down Features w/o ARM \  -\ -\ -\ -\ -\ -\ -\ $|$}
	\end{overpic}
	\caption{
		\textbf{Visualization of attention retaining with ARM}.
		The first row shows the multiple level intermediate features with (w/) our ARM,
		and the second row shows salient maps without (w/o) ARM.
		The prediction w/ ARM (second column, up) is more accurate than that w/o ARM (second column, down), since our ARM pays stronger attention to the co-salient regions.
		}
	\label{fig:ARM}
\end{figure}

To validate the effectiveness of our ARM, in \figref{fig:ARM}, we show the intermediate features in different levels of the decoder with ARM (1st row) and without ARM (2nd row).
We observe that, through GIM, both locate the co-salient object (\ie, Teddy bear) successfully, while our GICD w/o ARM is gradually interfered by other salient objects during upsampling and produces inaccurate detection results.
These results show that our ARM can effectively hold the attention on the co-salient objects in relevant images.

\subsection{Jigsaw Training Strategy}
\subsubsection{Strategy.}
\begin{figure}[t]
	\centering
	\begin{overpic}[width=0.85\columnwidth]{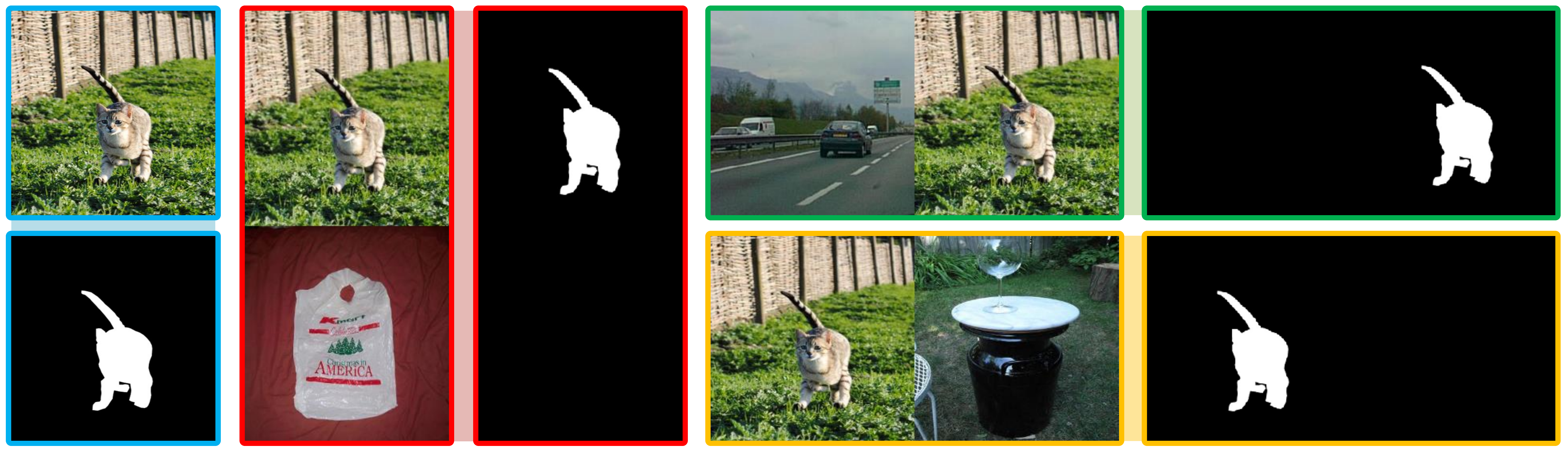}
	\end{overpic}
	\caption{
		\textbf{Demo of jigsaw training}.
		A sample (cat), together with the samples from other categories, constitute new jigsaws for training.
		}
	\label{fig:jigsaw}
\end{figure}

One important problem in Co-SOD task is that
current SOD datasets, \eg, DUTS~\cite{wang2017} and MSRA-B~\cite{liu2010learning}, are not suitable for the training of Co-SOD networks. 
The reasons are two-fold:
1) they do not have class information, so it is impossible to train models in groups;
2) most samples in them only contain one salient foreground object.
It is difficult to enable the network to distinguish the co-salient object(s) among multiple foreground objects.
Recent Co-SOD methods~\cite{wei2019deep,li2019detecting,wang2019robust} are trained on semantic segmentation datasets~\cite{lin2014microsoft}.
This suffers from two problems: 1) the label of such a semantic segmentation dataset is relatively rough, so the ability of recovery details of the trained network is not ideal, which cannot meet the accuracy requirements of downstream tasks;
2) the objects in such datasets are not necessarily salient. 
To alleviate these problems, we design a jigsaw strategy to transform SOD datasets into suitable training data for Co-SOD models:
Step 1: we employ a classifier~\cite{mahajan2018exploring} to classify every SOD dataset into multiple categories since it has no category information.
Step 2: we splice the samples of one category with the samples of other ones to form a new jigsaw, as shown in \figref{fig:jigsaw}.
This step is to ensure that an input image contains not only a co-salient foreground but also extraneous foreground objects.
Through the above steps, existing SOD datasets can be seamlessly utilized to train Co-SOD networks without additional pixel-level annotations.

\subsubsection{Loss function.}
Considering the most important goal for co-saliency detection is to find the position of the common foreground objects correctly,
we employ the soft intersection over union (IoU) loss~\cite{li2018interactive,qin2019basnet} for GICD, specifically,
\begin{equation}
\mathcal{L}\left(S, G\right) = 1 - \frac{\sum\limits_{c} S\left(c\right)G\left(c\right)}{\sum\limits_{c} \left[  S\left(c\right) + G\left(c\right) - S\left(c\right)G\left(c\right)  \right]},
\end{equation}
where $S$ is the prediction, and $G$ denotes ground-truth. $c$ represents each pixel position in the image. 
The loss function of our model can be expressed as
\begin{equation}
L_{total} = \sum_{n=1}^N \sum_{i=1}^4\mathcal{L}\left(S^i_n, G_n\right).
\end{equation}

\section{Proposed \textsl{CoCA} Dataset}
\label{coca}
\subsubsection{Construction guidelines.}
We construct our \textit{CoCA} dataset under four guidelines.
\textbf{G1}: each image should contain at least one extraneous foreground, excluding the co-salient object(s).
\textbf{G2}: in each image group, the co-salient objects are better to be different.
%
\textbf{G3}: the dataset needs to be misaligned with the categories of the common training set, to explore the ability of the model on handling unseen categories.
The guideline \textbf{G1} reflects whether or not the model can detect the co-salient objects, rather than only segmenting the foreground and background.
\textbf{G2} can evaluate whether the model is robust to the intra-group differences.
\textbf{G3} ensures that the model can be evaluated for its ability to detect co-salient objects from unknown categories robustly.

\subsubsection{Construction procedures.}
With the above guidelines, we collect images from pixabay\footnote{https://pixabay.com}.
We divide them into 80 categories, covering everyday indoor and outdoor scenes.
It is worth noting that these categories are outright staggered with Microsoft COCO~\cite{lin2014microsoft}, 
which is often used for the training of co-saliency models~\cite{wei2019deep,li2019detecting,wang2019robust}.
Most importantly, with manual screening, the images in our dataset include at least one extraneous salient object, excluding the co-salient object(s).
%
We provide four levels of annotations: class level, bounding box level, object level, and instance level.
The high-quality object-level annotations are applicable to the co-saliency detection task in this paper.
Different levels of annotations of our dataset corresponds to different tasks, such as co-localization~\cite{joulin2014efficient,tang2014co}, few-shot object segmentation~\cite{zhang2018sgone,zhang2019canet}, and instance co-segmentation~\cite{sun2017learning}.
%

\subsubsection{Dataset statistics.}
Our \textit{CoCA} dataset consists of 80 categories with 1295 images. 
As shown in \figref{fig:dataset},
these images are challenging in occlusion, clutter background, extraneous object interference \etc.
The number of images in each category is different, varying from 8 to 40.
This diversity is helpful in evaluating the ability of the model for different image set sizes.
The number of co-salient instances in an image is also diverse. 
336 images have more than two co-salient instances.
The diversity of the number of instances can help to evaluate the robustness of the model to multi-object scenarios.

\section{Experiments}
\label{sec:experiments}
\subsection{Implementation Details}
We train our GICD network on the training set of \textsl{DUTS}~\cite{wang2017} with our jigsaw training strategy.
These samples are classified into 291 groups, which contains 8250 images with removing the noisy samples. 
Each sample will be combined with others to form three jigsaws as supplementary samples, as shown in \figref{fig:jigsaw}; thus, the candidate training data is quadrupled.
In each training epoch, we randomly select at most 20 samples from each group.
The Adam optimizer~\cite{kingma2015adam} is used with an initial learning rate of 0.0001, $\beta_1=0.9$, and $\beta_2=0.99$.
The learning rate is divided by 10 at the $50$-th epoch.
We train our GICD for 100 epochs in total.
To accommodate the input images with our FPN backbone (VGG network), we resize them to $224\times224$ during the training and test stage, and the output saliency maps are resized back to the original size for evaluating.
Our GICD is implemented in PyTorch~\cite{pytorch2019paszke}, and runs at $\sim55$ FPS on an NVIDIA GeForce RTX 2080Ti.

\subsection{Evaluation Datasets and Metrics}
\textbf{Datasets.}
We employ two challenging datasets to evaluate the performance of various methods.
The first dataset is \textit{CoSal2015}~\cite{zhang2016CoSal}.
In some image groups, \eg~baseball, it is challenging in the interference of extraneous salient objects.
The other is our \textit{CoCA}, where most images possess more than one irrelevant salient objects besides the co-salient target.

\noindent
\textbf{Metrics.}
We employ five widely used metrics as suggested by \cite{zhang2018review,hsu2018unsupervised,zhang2019CSMG}: mean F-measure ($F_{\text{avg}}$)~\cite{achanta2009frequency}, maximum F-measure ($F_{\max}$)~\cite{borji2015SalObjBenchmark},  Precision-Recall (PR) curve, S-measure ($S_{\alpha}$)~\cite{fan2017structure}, and mean E-measure ($E_{\xi}$)~\cite{Fan2018Enhanced}.
%

\begin{table}[t!]
	\centering
		\footnotesize
	\renewcommand{\arraystretch}{1.25}
	\renewcommand{\tabcolsep}{1.6mm}
	\begin{tabular}{lr|cccc|ccc|c}
		\hline\toprule
		&          &    CBCD              &  GW         &   CSMG                 &   RCAN      &  BASNet  &  PoolNet  &    SCRN  &    GICD  \\
		&  Metric  & \cite{fu2013cluster} &  \cite{wei2017gw} &  \cite{zhang2019CSMG}  &  \cite{li2019detecting}  & \cite{qin2019basnet} &  \cite{liu2019poolnet}  &  \cite{wu2019SCRN}  &  \scriptsize{Ours}  \\
		\hline
		& $F_{\text{avg}}\uparrow$     & 0.378 &  0.639 &  0.721      &  0.670 & \tbb{0.778} & 0.768 & 0.755 &  \trb{0.835} \\
		& $F_{\max}\uparrow$  & 0.547 & 0.706 & 0.787 & 0.764 & \tbb{0.791} & 0.785 & 0.783 &  \trb{0.844} \\
		& $S_{\alpha}\uparrow$    & 0.550 &  0.744 &  0.776      &  0.779 & 0.822 & \tbb{0.823} & 0.817 &  \trb{0.844} \\
		\multirow{-4}{*}{\begin{sideways}CoSal2015\end{sideways}}
		& $E_{\xi}\uparrow$       & 0.516 &  0.727 &  0.763       &  0.742 & \tbb{0.841} & 0.836 & 0.822 &  \trb{0.883} \\
		\hline
		& $F_{\text{avg}}\uparrow$     & 0.230 &  0.358 &  0.390       &  0.360 & \tbb{0.398} & 0.394 & 0.394       & \trb{0.504} \\
		& $F_{\max}\uparrow$  & 0.313 & 0.408 & \tbb{0.499} & 0.422 & 0.408 & 0.404 & 0.413 &  \trb{0.513} \\
		& $S_{\alpha}\uparrow$    & 0.523 &  0.602 &  \tbb{0.627} &  0.616 & 0.592       & 0.602 & 0.612       & \trb{0.658} \\
		\multirow{-4}{*}{\begin{sideways}CoCA\end{sideways}}
		& $E_{\xi}\uparrow$       & 0.535 &  0.615 &  0.606       &  0.614 & 0.600       & 0.616 & \tbb{0.625} & \trb{0.701} \\
		\bottomrule
	\end{tabular}
	\caption{
		\textbf{Quantitative comparisons} of mean F-measure~\cite{achanta2009frequency} ($F_{\text{avg}}$),
		maximum F-measure~\cite{borji2015SalObjBenchmark} ($F_{\max}$),
		S-measure~\cite{fan2017structure} ($S_{\alpha}$),
		and mean E-measure~\cite{Fan2018Enhanced} ($E_{\xi}$) by our GICD and other methods on the \textit{CoSal2015}~\cite{zhang2016CoSal} 
		and \textit{CoCA} datasets.
		``$\uparrow$'' means that the higher the numerical value, the better the model performance.
	}
	\label{tab:Results}
\end{table}

\subsection{Comparison with State-of-the-Arts}
\subsubsection{Comparison  methods.}
We compare our GICD with seven state-of-the-art methods, including four Co-SOD method: RCAN~\cite{li2019detecting},
CSMG~\cite{zhang2019CSMG}, GW~\cite{wei2017gw}, and CBCD~\cite{fu2013cluster},
as well as three SOD mehtods: BASNet (ResNet-34)~\cite{qin2019basnet}, PoolNet (ResNet-50)~\cite{liu2019poolnet}, and SCRN (ResNet-50)~\cite{wu2019SCRN}.

\subsubsection{Quantitative evaluation.}
\begin{figure}[t]
	\centering
	\begin{overpic}[grid=false, width=1\columnwidth]{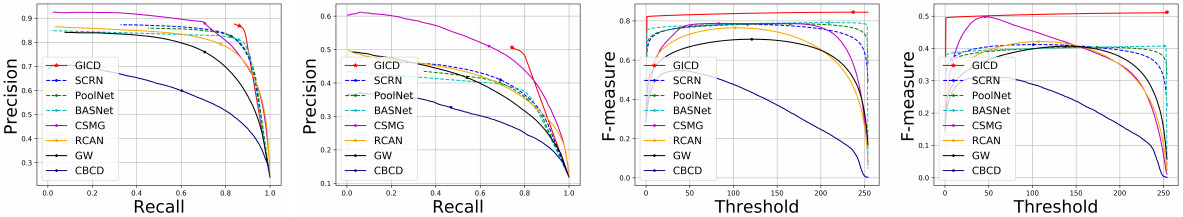}
	\put(6.6,19.3){\scriptsize{CoSal2015~\cite{zhang2016CoSal}}}
	\put(33,19.3){\scriptsize{CoCA (Ours)}}
	\put(57.4,19.3){\scriptsize{CoSal2015~\cite{zhang2016CoSal}}}
	\put(83,19.3){\scriptsize{CoCA (Ours)}}
	\end{overpic}
	\caption{
		\textbf{Precision-Recall (PR) and F-measure curves} of our GICD and seven state-of-the-art methods on the \textit{CoSal2015} and \textit{CoCA} datasets. 
		The node on each PR curve denotes the
		precision and recall value used for calculating maximum F-measure.}
	\label{fig:PR_Fm}
\end{figure}
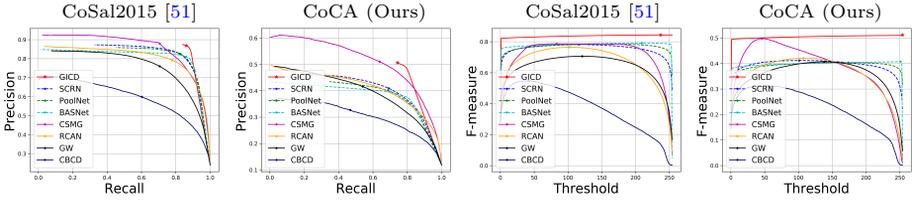

In the \tabref{tab:Results},
we illustrate the quantitative results of our GICD and other state-of-the-art methods on the \textit{CoSal2015} and our \textit{CoCA} datasets.
As can be seen, 
our GICD achieves better performance.
The results show some interesting phenomena.
On the \textit{CoSal2015}, the SOD methods outperform most Co-SOD methods except GICD.
The reason is that a large part of the images in \textit{CoSal2015} have only one salient object, 
which can be solved by SOD algorithms.
The advantages of Co-SOD algorithms cannot be fully reflected on this data, 
and these detail-oriented SOD methods easily surpass their performance.
However, in our newly proposed \textit{CoCA} dataset, 
this phenomenon is no longer obvious, because the salient objects in an image contain many objects that are not co-salient.
This is why our \textit{CoCA} dataset is more suitable for evaluating Co-SOD algorithms.
Nevertheless, our GICD still surpasses the SOD methods on \textit{CoSal2015}.
It brings 11.4\% improvement in terms of mean F-meansure compared with the best Co-SOD method,
5.7\% improvement compared with the SOD method.
In our \textit{CoCA} dataset,
GICD brings 3.1\% improvement in terms of S-measure compared with the best Co-SOD method,
4.6\% improvement compared with the SOD method.
Seen from \figref{fig:PR_Fm}, our method also outperforms other methods on the PR curve and F-measure curve. 
The trend of the curves demonstrates
our method is less affected by the threshold, because it predicts the result with high confidence.
This can avoid the problem of how to select the appropriate threshold in the subsequent practical applications.

\subsubsection{Qualitative results.}
\begin{figure}[t]
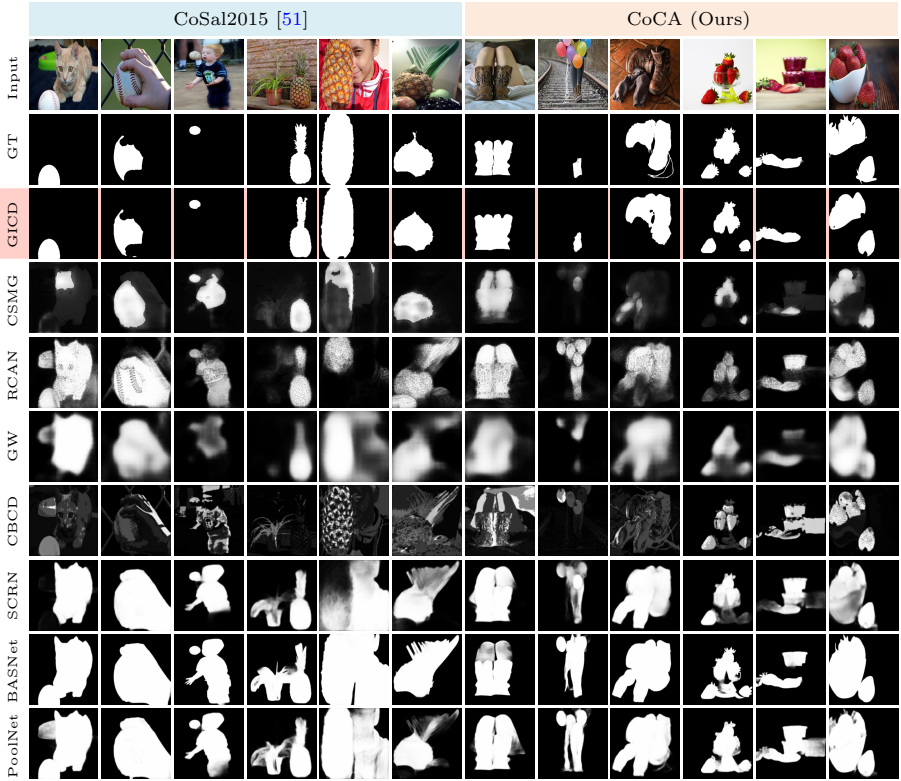

	\centering
	\begin{overpic}[grid=false, width=1\columnwidth]{./Images/Salmaps.pdf}
	\put(19.8,82){\scriptsize{CoSal2015~\cite{zhang2016CoSal}}}
	\put(69.5,82){\scriptsize{CoCA (Ours)}}
	\put(1.7,74){\tiny{\rotatebox{90}{Input}}}
	\put(1.7,67.5){\tiny{\rotatebox{90}{GT}}}
	\put(1.7,58){\tiny{\rotatebox{90}{GICD}}}
	\put(1.7,49.5){\tiny{\rotatebox{90}{CSMG}}}
	\put(1.7,41.5){\tiny{\rotatebox{90}{RCAN}}}
	\put(1.7,35){\tiny{\rotatebox{90}{GW}}}
	\put(1.7,25.8){\tiny{\rotatebox{90}{CBCD}}}
	\put(1.7,18){\tiny{\rotatebox{90}{SCRN}}}
	\put(1.7,9){\tiny{\rotatebox{90}{BASNet}}}
	\put(1.7,1){\tiny{\rotatebox{90}{PoolNet}}}
	\end{overpic}
	\caption{\textbf{Visual comparison} of our GICD with 7 state-of-the-arts (4 Co-SOD methods and 3 SOD methods) on the \textit{CoSal2015}~\cite{zhang2016CoSal} and our \textit{CoCA} datasets.}
	\label{fig:salmaps}
\end{figure}

In \figref{fig:salmaps}, we show some saliency maps produced by GICD and other compared methods for intuitive comparison.
The samples we illustrate are challenging because the salient objects in each input include not only co-salient object(s)
but also interference from other extraneous foreground(s).
This is also reflected in the prediction results of SOD algorithms, 
which over segmented many unrelated regions.
From the overall results, our GICD has high confidence in the prediction maps, even at the edge, 
while most of the other methods suffer from uncertain regions.
Back to specific examples,
baseball is the most challenging subset of the \textit{CoSal2015}~\cite{zhang2016CoSal}, 
because it varies greatly in size across images and is interfered by other salient objects.
Results show that our method successfully handles the tiny size and the occlusions.
In \textit{CoCA}, boots class faces the interference of background color, and strawberry class has multiple segmentation targets.
Nevertheless, GICD locates the target object accurately.

\subsection{Ablation Study}
\label{sec:abl}
In order to explore the contribution and mechanism of gradient inducing module (GIM), attention retaining module (ARM), and jigsaw training (JT) to our GICD network, 
we evaluated all possible combinations of the three candidates.
Note that, 
these three candidates are interdependent and are not recommended to be used alone.
As shown in \tabref{tab:ABL},
``A'' is the baseline without JT, GIM, and ARM.
It is actually a SOD model because it does not take into account any relationship between images.

\subsubsection{Effectiveness of GIM.}
GIM is the core module of our GICD.
With GIM, the variants of ``C'', ``E'', ``G'', and our GICD can be regarded as a Co-SOD network; while, without GIM, the variants of ``A'', ``B'', ``D'', and ``F'' have limited ability on discovering co-salient objects.
By directly applying GIM, the variant ``C'' shifts the attention to co-salient objects in the high-level features, and achieves a certain performance improvement compared to the baseline.
However, in this case, the training set is ill-posed for co-saliency detection task without JT, and the attention will be disturbed during the decoding without the help of ARM.
These factors limit its performance.
By introducing JT or ARM (variants ``E'', ``G'' and our GICD), the effect of GIM is further enhanced.

\subsubsection{Effectiveness of ARM.}
ARM plays a role in retaining high-level prediction information during the top-down decoding.
As shown in variant ``D'', using ARM alone does not improve Co-SOD performance.
The reason is that, without inducing by GIM, the prediction in high-level is actually the salient objects rather than co-salient objects.
When cooperating with GIM in variant ``G'', although trained on ill-posed data, 
it still compulsorily keeps the inducing information of GIM to an extend; thus, ``G'' achieves significantly better performance than the variant ``C''.
``E'' is a variant with our GIM and ARM modules.
As shown in \figref{fig:ARM}, without ARM, it is easy to be interfered by irrelevant foreground when recovering object details.
Therefore, its performance is inferior to our GICD.

\subsubsection{Effectiveness of JT.}
The jigsaw training (JT) helps turn SOD datasets into Co-SOD ones, and serves as a useful strategy for training Co-SOD networks.
In \tabref{tab:ABL}, without GIM, the variant model ``B'' and ``F'' are SOD models, not Co-SOD ones.
Since no interactive cue between images is considered,
a SOD model trained on Co-SOD dataset is unable to discover group-wise co-salient connections, and the generated JT labels will bring meaningless predictions in this ill-posed scene;
therefore, the JT does not work in these cases.
When working with GIM in variant ``E'',  JT improves the effect in the challenging \textit{CoCA} dataset.
Similarly, this improvement can also be seen through the comparison between our GICD and variant ``G''.

In summary, our three contributions of the GIM, ARM, and JT candidates are mutually reinforced for better co-saliency detection performance, as validated through comprehensive experiments.

\begin{table}[t]
	\centering
	\scriptsize
	\renewcommand{\arraystretch}{1.4}
	\renewcommand{\tabcolsep}{1.6mm}
	\begin{tabular}{c|ccc|cccc|cccc}
		\hline\toprule
		\multirow{2}{*}{Variant} & \multicolumn{3}{c|}{Candidate}   & \multicolumn{4}{c|}{CoCA} & \multicolumn{4}{c}{CoSal2015~\cite{zhang2016CoSal}}   \\
		        & JT         & GIM          & ARM    & $F_{\text{avg}}\uparrow$ & $F_{\max}\uparrow$ & $S_{\alpha}\uparrow$ & $E_{\xi}\uparrow$ &  $F_{\text{avg}}\uparrow$ & $F_{\max}\uparrow$ & $S_{\alpha}\uparrow$ & $E_{\xi}\uparrow$  \\
		\hline
		A  &            &             &               & 0.420 & 0.430 & 0.601 & 0.627 & 0.788 & 0.800 & 0.818 & 0.852  \\
	    B  &\checkmark  &             &               & 0.424 & 0.430 & 0.602 & 0.655 & 0.750 & 0.759 & 0.782 & 0.821  \\
		C  &            &  \checkmark &               & 0.446 & 0.462 & 0.618 & 0.643 & 0.809 & 0.824 & 0.833 & 0.868  \\
		D  &            &             & \checkmark    & 0.429 & 0.437 & 0.607 & 0.628 & 0.800 & 0.809 & 0.829 & 0.860  \\
		E  &\checkmark  &  \checkmark &               & 0.470 & 0.478 & 0.631 & 0.689 & 0.795 & 0.803 & 0.808 & 0.850  \\
		F  & \checkmark &             & \checkmark    & 0.436 & 0.442 & 0.612 & 0.654 & 0.762 & 0.770 & 0.795 & 0.832  \\
		G  &            &  \checkmark & \checkmark    & 0.471 & 0.480 & 0.636 & 0.667 & 0.826 & 0.835 & 0.845 & 0.879  \\
		\hline
		GICD  & \checkmark &  \checkmark & \checkmark & 0.504 & 0.513 & 0.658 & 0.701 & 0.835 & 0.844 & 0.844 & 0.883  \\
		\bottomrule
	\end{tabular}
	\caption{
		\textbf{Ablation study} of the proposed GICD on the \textit{CoCA} and \textit{CoSal2015} datasets.
		The candidates are jigsaw training (JT), gradient inducing module (GIM), and attention retaining module (ARM).
		Note that, the variants ``A'', ``B'', ``D'', and ``F'', without GIM, are actually SOD models rather than Co-SOD models.
		The experiments reflect the interaction mechanism of our three contributions.
	}
	\label{tab:ABL}
\end{table}

\section{Conclusions}
In this paper, inspired by the mechanism of how human behaves on the Co-SOD task, we proposed an end-to-end Gradient-Induced Co-saliency Detection (GICD) method.
In GICD, the gradient information, which highlights the discrimination of features, is generated from the comparison between single and consensus representations.
Induced by the gradient, GICD pays more attention to discriminative convolutional kernels, enabling our model to locate the co-salient regions.
Due to the lack of Co-SOD training data, we designed a novel jigsaw training strategy, with which we trained Co-SOD models on a general SOD dataset without extra pixel-level annotations.
In addition, we constructed a challenging \textit{CoCA} dataset for Co-SOD evaluation, to prosper the subsequent research on exploring real-world Co-SOD scenarios.

\paragraph{\textbf{Acknowledgements.}}
Ming-Ming Cheng is the corresponding author.
Zhao Zhang and Wenda Jin are the joint first authors.
This research was supported by Major Project for New
Generation of AI under Grant No. 2018AAA0100400,
NSFC (61922046), Tianjin Natural Science Foundation (18ZXZNGX00110),
and the Fundamental Research Funds for the Central Universities,
Nankai University (63201169).

\bibliographystyle{splncs04}
\bibliography{egbib}
\end{document}